%% file: main.tex
\pdfoutput=1
\documentclass[11pt]{article}

\usepackage[]{ACL2023}

\usepackage{times}
\usepackage{latexsym}

\usepackage[T1]{fontenc}
\usepackage[utf8]{inputenc}

\usepackage{microtype}
\usepackage{subfigure}
\usepackage{graphicx}
\usepackage{multirow}
\usepackage{booktabs}
\usepackage{tabularx}
\usepackage{bm}
\usepackage{amsmath}
\usepackage{inconsolata}
\title{PAL: Persona-Augmented Emotional Support Conversation Generation}

\author{
Jiale Cheng\thanks{\ \ Equal contribution.} , Sahand Sabour\footnotemark[1] , Hao Sun , Zhuang Chen , Minlie Huang\thanks{\ \ Corresponding author.} \\
\small{The CoAI group, DCST; Institute for Artificial Intelligence; State Key Lab of Intelligent Technology and Systems;}\\
\small{Beijing National Research Center for Information Science and Technology;} 
\small{Tsinghua University, Beijing 100084, China.}\\ \small{\texttt{\{chengjl19,sm22,h-sun20,zhchen-nlp\}@mails.tsinghua.edu.cn,}}
\small{\texttt{aihuang@tsinghua.edu.cn}} \\
}

\begin{document}
\maketitle
\begin{abstract}
    \input{sections/abstract}

\end{abstract}

\section{Introduction}
    \input{sections/introduction}

\section{Related Work}
    \input{sections/related_work}

\section{Persona-Augmented Emotional Support}
    \input{sections/method}

\section{Experiments}
    \input{sections/experiment}

\section{Case Study}
    \input{sections/case_study}

\section{Conclusion}
    \input{sections/conclusion}

\section*{Limitations}
    \input{sections/limitations}

\section*{Ethical Considerations}
    \input{sections/ethics}

% \section*{Acknowledgements}
    % \input{sections/acknowledgement.tex}
% Entries for the entire Anthology, followed by custom entries
\bibliography{anthology,custom}
\bibliographystyle{acl_natbib}

\appendix
    \input{sections/appendix}

\end{document}

%% file: sections/abstract.tex
Due to the lack of human resources for mental health support, there is an increasing demand for employing conversational agents for support.
Recent work has demonstrated the effectiveness of dialogue models in providing emotional support.
As previous studies have demonstrated that seekers' persona is an important factor for effective support, we investigate whether there are benefits to modeling such information in dialogue models for support. 
In this paper, our empirical analysis verifies that persona has an important impact on emotional support. 
Therefore, we propose a framework for dynamically inferring and modeling seekers' persona. 
We first train a model for inferring the seeker's persona from the conversation history. 
Accordingly, we propose PAL, a model that leverages persona information and, in conjunction with our strategy-based controllable generation method, provides personalized emotional support. 
Automatic and manual evaluations demonstrate that PAL achieves state-of-the-art results, outperforming the baselines on the studied benchmark. 
Our code and data are publicly available at \url{https://github.com/chengjl19/PAL}.

%% file: sections/introduction.tex
% Importance of Support
A growing number of people are experiencing mental health issues, particularly during the Covid-19 pandemic \cite{hossain2020epidemiology, talevi2020mental, cullen2020mental, kumar2021covid}, and more and more people are seeking mental health support.
The high costs and limited availability of support provided by professional mental health supporters or counselors \cite{AlanEKazdin2011RebootingPR, MarkOlfson2016BuildingTM, denecke2020mental, EvanPeterson2021WisconsinMH} have highlighted the importance of employing conversational agents and chatbots for automating this task \cite{cameron2018assessing, daley2020preliminary, denecke2020mental, kraus2021towards}.

\begin{figure}[ht]
\centering
\includegraphics[width=\linewidth]{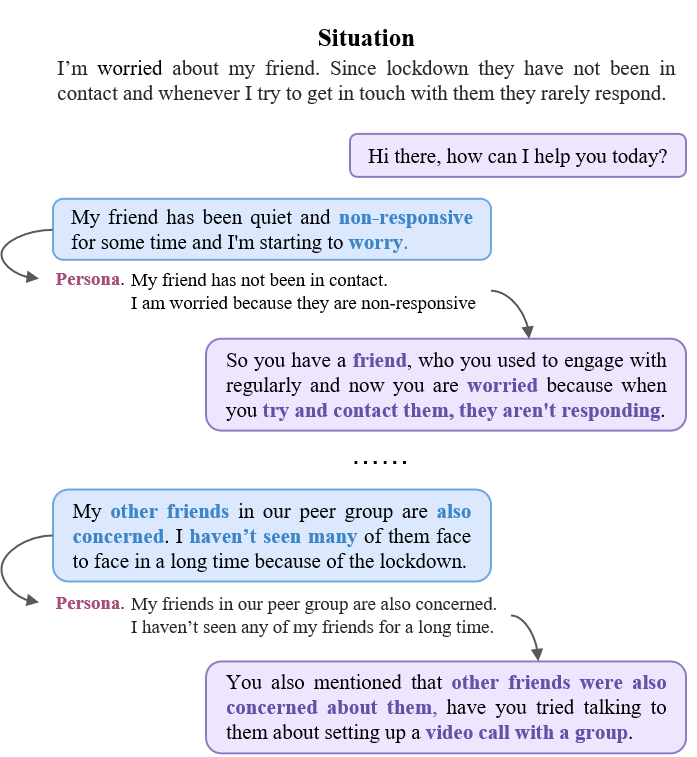} 
\caption{An example from the ESConv dataset, in which the trained supporter extracts key information about the seeker's persona and leverages this information to provide effective emotional support.}
\label{fig:intro} 
\end{figure}

% What is Emotional Support?
Towards this end, \citet{liu2021towards} pioneered the task of emotional support conversation generation to reduce users' emotional distress and improve their mood using language models. 
They collected ESConv, a high-quality crowd-sourced dataset of conversations (with annotated helping strategies) between support seekers and trained emotional supporters, and demonstrated that training large pre-trained dialogue models on this dataset enabled these models to provide effective support.
\citet{tu2022misc} proposed to leverage commonsense knowledge and implemented hybrid strategies to improve the performance of dialogue models in this task.
Similarly, \citet{peng2022control} also suggested using commonsense knowledge for this task and further proposed a global-to-local graph network to model local and global hierarchical relationships. More recently, \citet{cheng2022improving} proposed look-ahead strategy planning to select strategies that are more effective for long-turn interactions.

% What is missing?
Although previous studies have considered relevant psychological theories and factors, such as commonsense reasoning, they neglect information regarding the users' persona. Persona, which can be considered as an outward expression of personality \cite{leary2011personality} in Psychology, is also closely related to empathy \cite{richendoller1994exploring, costa2014associations}, anxiety \cite{smrdu2021covid}, frustration \cite{Jeronimus2017}, mental health \cite{michinov2021stay} and distress \cite{liu2018relationship}, all of which are essential concepts in psychological scenarios. 
Effective emotional support benefits from an adequate understanding of the support seeker's personality, as shown by research on person-centered therapy \cite{rogers2013client}, while more specific and persona-related words lead to a long-term rapport with the user \cite{campos2018challenges}. 
Thus, the inability to actively combine persona information and conversations prevents users from developing such rapport with the system \cite{xu2022long}, which is not desirable for emotional support.
Therefore, it is intuitive to explore seekers' personas and build systems for providing personalized emotional support. 

In this paper, we propose \textbf{P}ersona-\textbf{A}ugmented Emotiona\textbf{L} Support (\textbf{PAL}), a conversational model that learns to dynamically leverage seekers' personas to generate more informative and personalized responses for effective emotional support. 
%%%这里有点相悖的地方：既是user persona，又是推导出来的；那如何保证是否真的符合用户persona
To more closely match realistic scenarios (no prior knowledge of the user's persona) and retain important user information from earlier conversation rounds, we first extract persona information about the seeker based on the conversation history and design an attention mechanism to enhance the understanding of the seeker. Furthermore, we propose a strategy-based controllable generation method to actively incorporate persona information in responses for a better 
rapport with the user.
We conduct our experiments on the ESConv dataset \cite{liu2021towards}. Our results demonstrate that PAL outperforms the baselines in automatic and manual evaluations, providing more personalized and effective emotional support. We summarize our contributions as follows:
% TODO精简一下contribution
\begin{itemize}
    \item To the best of our knowledge, our work is the first approach that proposes to leverage persona information for emotional support. 
    
    \item We propose a model for dynamically extracting and modeling seekers' persona information and a strategy-based decoding approach for controllable generations.
    
    \item Our analysis of the relationship between the degree of individuality and the effect of emotional support, in addition to the conducted experiments on the ESConv dataset and comparisons with the baselines, highlights the necessity and effectiveness of modeling and leveraging seekers' persona information.
\end{itemize}

%% file: sections/related_work.tex
\subsection{Persona in Conversation Generation}
There are extensive studies on leveraging persona information in dialogue \cite{10.1145/3383123}. 
However, it's important to note that the definition of persona in this context differs from its definition in Psychology. In dialogue systems, persona refers to the user's characteristics, preferences, and contextual information, which are incorporated to enhance the system's understanding and generation capabilities.
\citet{li2016persona} proposed using persona embeddings to model background information, such as the users' speaking style, which improved speaker consistency in conversations. 
However, as stated by \citet{xu2022long}, this approach is less interpretable. 
Therefore, several approaches to directly and naturally integrate persona information into the conversation were proposed \cite{zhang2018personalizing, wolf2019transfertransfo, liu2020you, yang2021generating}.

% Annotated persona
\citet{zhang2018personalizing} collected \textsc{persona-chat}, a high-quality dataset with annotated personas for conversations collected by crowd-sourcing workers.
This dataset has been widely used to further explore personalized conversation models and how persona could benefit response generation in conversations \cite{wolf2019transfertransfo, liu2020you, yang2021generating}.
% TODO 这里需要修改一下
However, it is relatively difficult to implement users' personas in real-world applications, as requiring users to provide information regarding their personas prior to conversations is impractical and unnatural. 

% Persona extraction and generation
\citet{xu2022long} addressed this problem by training classifiers that determine whether sentences in the conversation history include persona information. Accordingly, they store such sentences and leverage them to generate responses. However, in many cases, users do not explicitly express persona information in the conversation, which often requires a certain level of reasoning. 
For instance, a user may say, "My friend likes to play Frisbee, so do I", which does not contain any explicit persona information, but one could infer that the user likes to play Frisbee. 
In this work, we aim to infer possible persona information from the conversation history to assist our model in better understanding the user.

\subsection{Emotional Support}
In recent years, an increasing number of approaches have focused on emotional and empathetic response generation \cite{zhou2018emotional, zhong2020towards, kim2021perspective, gao2021improving, zheng2021comae, sabour2022cem}.
However, although such concepts are essential, they are insufficient for providing effective support as this task requires tackling the user's problem via various appropriate support strategies while exploring and understanding their mood and situation \cite{liu2021towards, AugESC}.
Therefore, \citet{liu2021towards} proposed the task of Emotional Support Conversation Generation and created a set of high-quality conversations between trained crowd-sourcing workers.
Their work demonstrated that training widely-used dialogue models, such as Blenderbot \cite{roller2021recipes}, on their collected dataset enabled such models to provide effective emotional support.
Following their work, \citet{tu2022misc} proposed leveraging external commonsense knowledge to better understand the users' emotions and suggested using a mixture of strategies for response generation.
\citet{peng2022control} implemented a hierarchical graph network to model the associations between global causes and local intentions within the conversation. 
\citet{cheng2022improving} proposed multi-turn strategy planning to assist in choosing strategies that are long-term beneficial.
However, existing work has not explored the effects of dynamically modeling users' persona information in this task, which we hypothesize improves models' emotional support ability and enables more personalized support.

%% file: sections/method.tex
\begin{figure}[htbp]
\centering
\includegraphics[width=\linewidth]{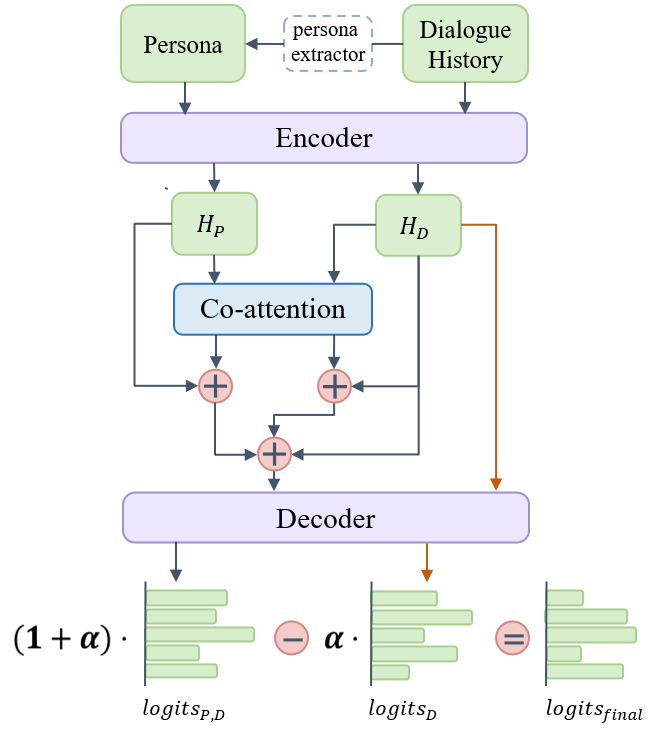} 
\caption{The overall structure of Persona-Augmented Emotional Support (PAL). We extract the seeker's persona from the dialogue history and then use a controllable generation method to generate the response. $\alpha$ is a tunable hyperparameter.}
\label{fig: model} 
\end{figure}

Figure \ref{fig: model} shows the overall flow of our approach. 
We first infer the seeker's persona information from the conversation history and then leverage the inferred information to generate a response. 
Our approach is comprised of three major components: The persona extractor for inferring the seeker's persona information (\S\ref{perosna extractor}); The response generator that leverages the inferred persona information and generates the response distribution (\S\ref{model architecture}); A strategy-based controllable decoding method for generating appropriate responses (\S\ref{decoding strategy}).

% \begin{figure}[htbp]
% \centering
% \includegraphics[width=\linewidth]{fig/model.png} 
% \caption{The Structure of Our Model.}
% \label{fig: model} 
% \end{figure}

\subsection{Problem Formulation} \label{sec:problem}
For inferring users' personas, we leveraged the \textsc{persona-chat} dataset \cite{zhang2018personalizing}, a high-quality collection of conversations between crowd-sourcing workers assigned with a set of pre-defined persona sentences.
Assume that a conversation between two speakers A and B is represented as $ D=\{u_1^A, u_1^B, u_2^A, u_2^B, \ldots, u_n^A, u_n^B\} $, where $u_i^A$ and $u_i^B$ represent the respective utterances of each speaker in the conversation, and $n$ indicates the number of utterances.
Accordingly, assume that each speaker has a set of persona information $ P_A = \{p_1^A, \ldots, p_{m_A}^A\} $ and $ P_B = \{p_1^B, \ldots, p_{m_B}^B\} $, where $p_i^A$ and $p_i^B$ represent the persona sentences for each speaker, respectively.
Our pioneer task is to infer a speaker's persona information based on their utterances in the conversation (e.g., inferring $P_A$ from $U_A=\{u_1^A,  u_2^A, \ldots, u_n^A\}$). 

As mentioned, we adopt the ESConv dataset \cite{liu2021towards} to train our model for providing emotional support.
Assume that a conversation between a support seeker A and supporter B at the $t^{\text{th}}$ turn of the conversation is $ D=\{u_1^A, u_1^B, u_2^A, u_2^B, \ldots, u_t^A\}$, where $u_i^A$ and $u_i^B$ represent the utterances of the seeker and the supporter, respectively.
Our task is two-fold: First, we infer the seeker's persona information $P_A$ from their utterances $U_A=\{u_1^A,  u_2^A, \ldots, u_t^A\}$.
Accordingly, we leverage the inferred information $P_A$ and conversation history $D$ to generate an appropriate supportive response $u_t^B$.

\subsection{Persona Extractor} \label{perosna extractor}
As previously stated, it is beneficial and essential to study the effects of leveraging persona information in the emotional support task. 
As predicting the seeker's persona information before the conversation is impractical, inferring such information from their utterances is necessary.

\begin{table*}[!htbp]
\begin{tabular}
	{
    >{
    \arraybackslash}p{
    0.57\textwidth}
	>{
    \arraybackslash}p{
    0.33\textwidth}}
\toprule
\textbf{Conversation} & \textbf{Persona} \\ \midrule
\textbf{Seeker:}    Hello & --- \\
\textbf{Supporter:} Hi there! How may I support you today? & --- \\  
\textbf{Seeker:} I'm just feeling anxious about my job's future. A lot of my colleagues are having trouble getting their licenses because of covid which means we won't be able to work. & \multirow{3}{*}{I am worried about my job's future.}\\
\textbf{Supporter:} That must be hard. COVID has turned our world upside down! What type of occupation are you in? & \multirow{2}{*}{I am worried about my job's future.} \\
\textbf{Seeker:}   I'm studying to be a pharmacist.&
I am worried about my job's future. \\
& I'm studying to be a pharmacist. \\
\bottomrule
\end{tabular}
\caption{An example conversation from PESConv. This conversation contains 5 utterances, where "---" indicates that no persona information was found. Once detected, new inferences are added to the seekers' persona.}
\label{persona annotation example}
\end{table*}

Based on the problem formulation in \S\ref{sec:problem}, we fine-tune a bart-large-cnn\footnote{\url{https://huggingface.co/facebook/bart-large-cnn}} to augment the ESConv \cite{liu2021towards} dataset with the inferred persona information annotations for each turn of the conversations. More details can be found in Appendix \ref{appendix: persona extractor}.
Since the initial utterances of this dataset generally contain greetings, we annotate the persona information starting from the third utterance of the conversation.  Table \ref{persona annotation example} shows an example of such annotations. 
We refer to this dataset with the additional annotations as Personalized Emotional Support Conversation (PESConv).

%To verify that modeling persona information in emotional support task is indeed important, we perform an analysis using our PESConv.
We analyze PESConv to confirm that modeling persona is essential for emotional support.
In the original ESConv dataset, workers score conversations based on the supporter's empathy level, the relevance between the conversation topic and the supporter's responses, and the intensity of the seeker's emotion. 
For each of these three aspects, we calculate the average cosine similarity between the responses and persona information in a conversation to examine how closely the responses and persona information are related. 

\begin{figure*}[ht]
\centering
\subfigure[Empathy]{
\begin{minipage}[t]{0.3\linewidth}
\centering
\includegraphics[width=2in, height=1.5in]{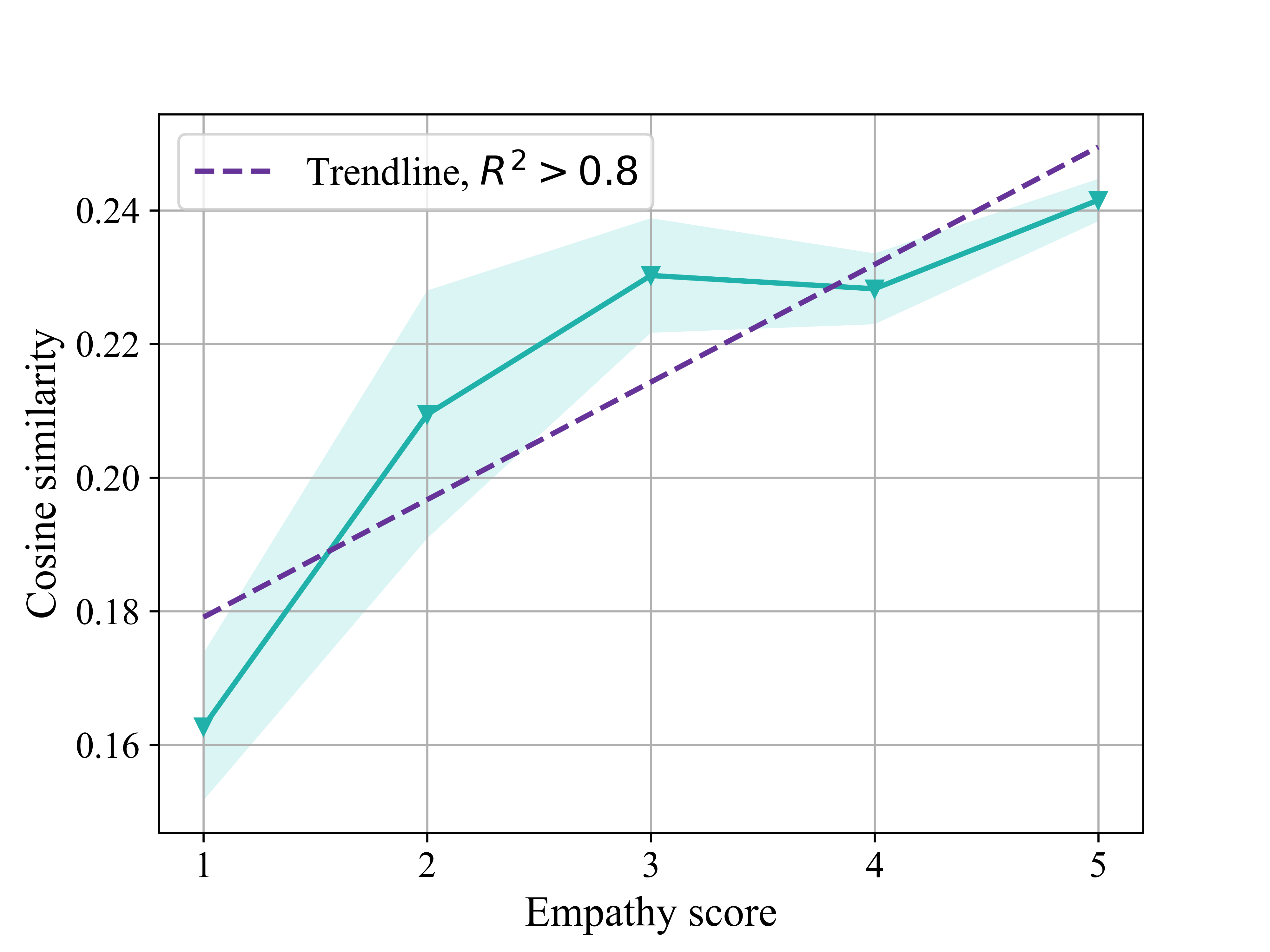}
\end{minipage}
\label{fig:simcse analysis_emp}
}
\subfigure[Relevance]{
\begin{minipage}[t]{0.3\linewidth}
\centering
\includegraphics[width=2in, height=1.5in]{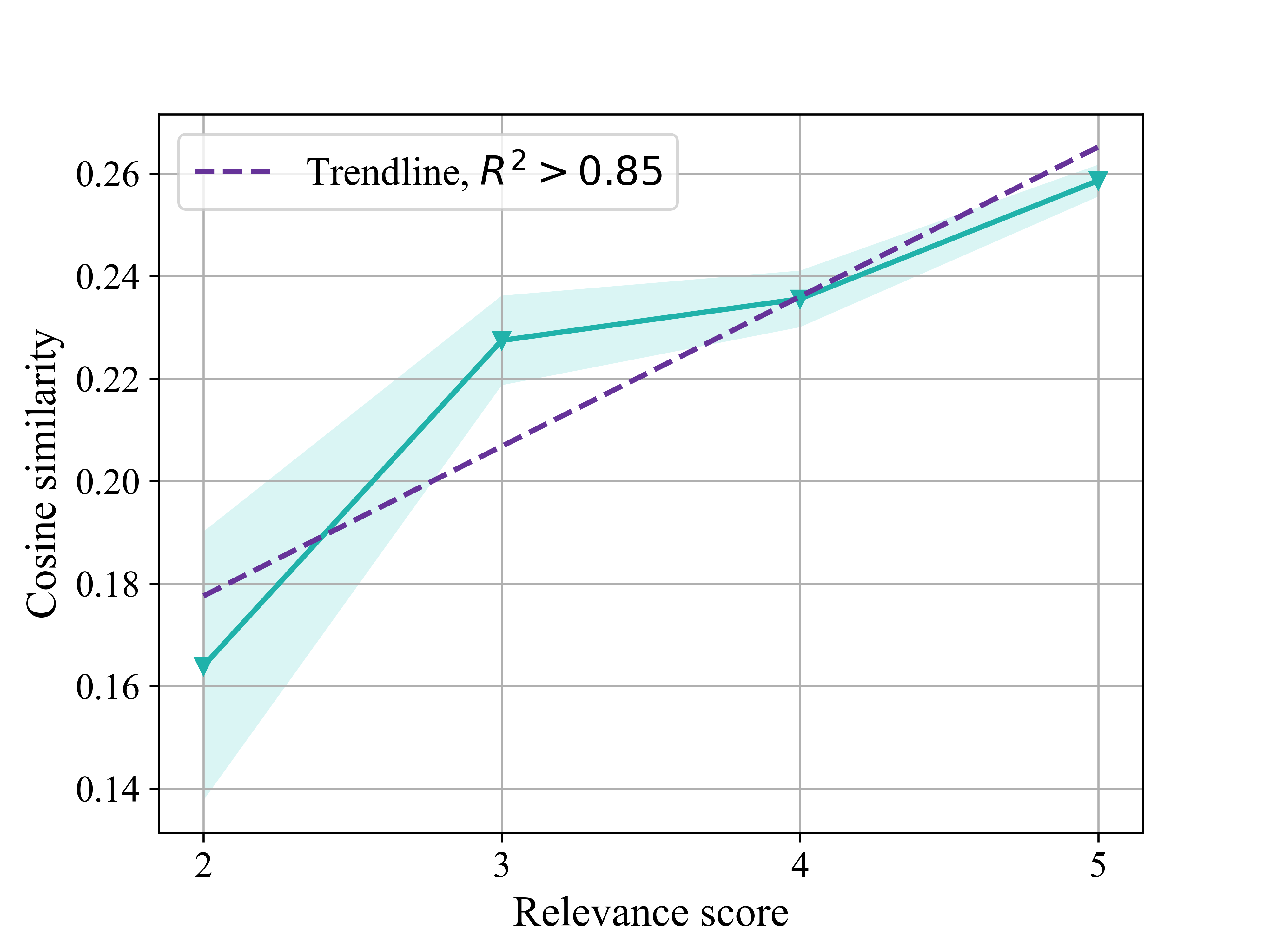}
\end{minipage}
}
\subfigure[Decrease in Emotional Intensity]{
\begin{minipage}[t]{0.3\linewidth}
\centering
\includegraphics[width=2in, height=1.5in]{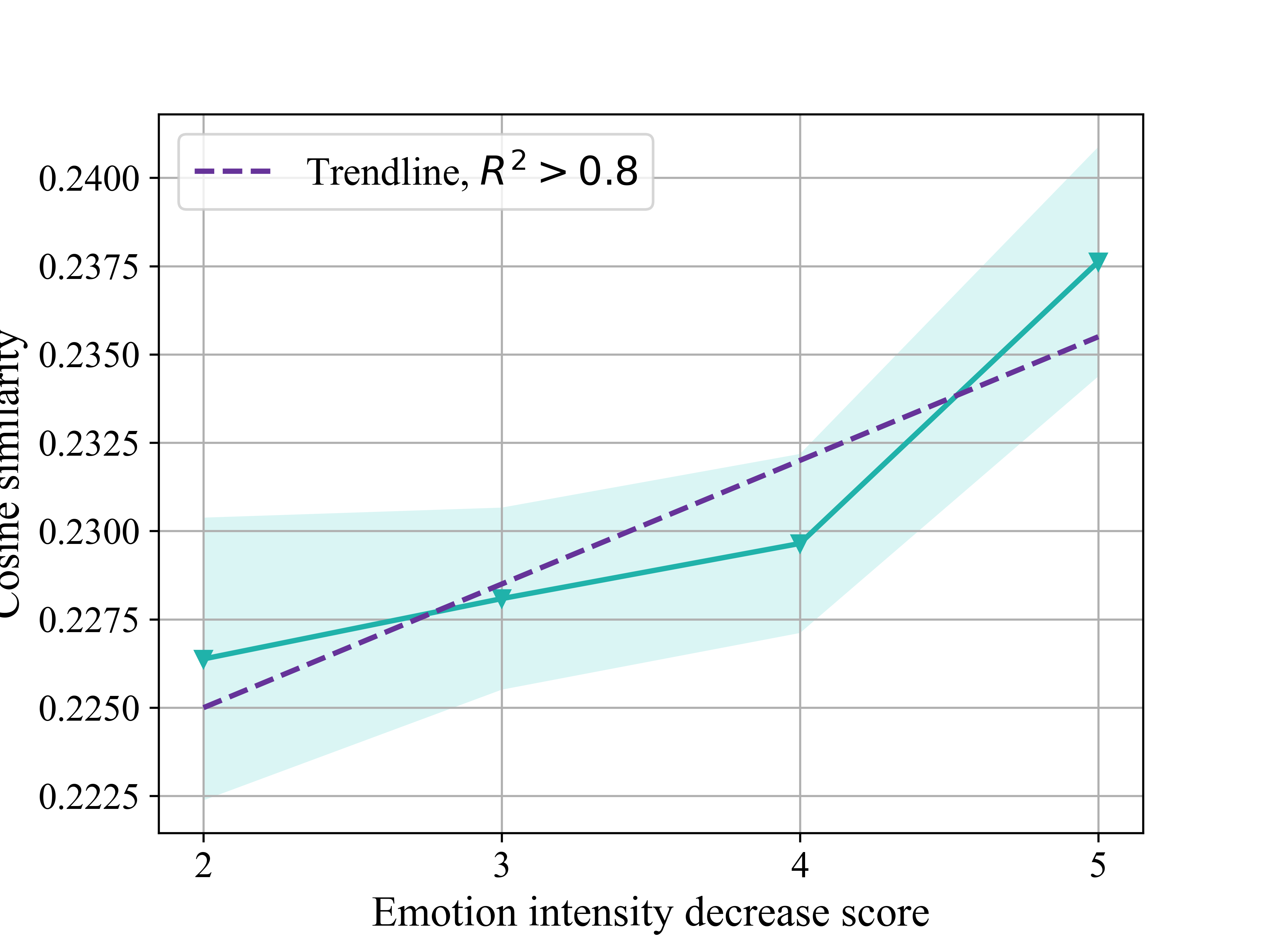}
\end{minipage}
}
\centering
\caption{The relationship between the empathy score, relevance score, emotion intensity decrease score, and the similarity between supporters' responses and persona information using SimCSE. We can observe that in general, more similarity leads to higher scores. In addition, we display the trend line and the coefficient of determination.}
\label{fig:simcse analysis}
\end{figure*}

For this task, we leverage SimCSE~\cite{gao2021simcse}, a sentence embedding model trained with a contrastive learning approach, to obtain vector representations for the sentences in PESConv. 
As illustrated in Figure \ref{fig:simcse analysis}, clearer and more appropriate mentions of the seekers' persona in the supporters' response lead to higher values for the studied aspects (i.e. higher empathy, more relevance, and a larger decrease in emotional intensity). 
Therefore, we believe this further highlights the necessity of modeling persona information in providing effective emotional support.
Moreover, we use fastText~\cite{joulin2017bag}, which represents sentences as averaged word embeddings, and the results (Appendix \ref{appendix: analysis}) demonstrate similar findings.

\subsection{Modeling Seekers' Persona} \label{model architecture}
As illustrated in Figure \ref{fig: model}, our model considers persona information as the model input in addition to the dialogue history. 
Formally, we use Transformer \cite{vaswani2017attention} encoders to obtain the inputs' hidden representations, which can be expressed as
\begin{equation}
\begin{aligned}
    &\bm{H}_D = \mathbf{Enc}(u_1, {\rm SEP}, u_2, \ldots, u_n ) \\
    &\bm{H}_P = \mathbf{Enc}(p_1, {\rm SEP}, p_2, \ldots, p_m ), 
\end{aligned}
\end{equation}
where $\mathbf{Enc}$ is the Transformer encoder, and $m$ and $n$ represent the number of persona sentences and conversation utterances, respectively. 
% Similar to \citet{devlin2019bert}, we add the special tokens $ \rm CLS $ and $ \rm SEP $ for classification (i.e., response strategy classification for our task) and sentence separation, respectively.
We use the special token $\rm SEP$ for sentence separation.

To highlight the context related to seekers' persona, we calculate an extra attention $\bm{Z}_D$ on $ \bm{H}_D $ and obtain a new hidden representation $\hat{\bm{H}}_D$ for dialogue history as follows:
\begin{equation}
\begin{aligned}
    &\bm{Z}_D = {\rm softmax}(\bm{H}_D \cdot \bm{H}_P^T) \cdot \bm{H}_P \\
    &\hat{\bm{H}}_D = {\rm LN}(\bm{H}_D + \bm{Z}_D)
\end{aligned}
\end{equation}
where $ \rm LN $ stands for the LayerNorm operation \cite{ba2016layer}.
Similarly, to promote persona sentences that are more aligned with the provided context, we obtain $\hat{\bm{H}}_P$ by
\begin{equation}
\begin{aligned}
    &\bm{Z}_P = {\rm softmax}(\bm{H}_P \cdot \bm{H}_D^T) \cdot \bm{H}_D \\
    &\hat{\bm{H}}_P = {\rm LN}(\bm{H}_P + \bm{Z}_P).
\end{aligned}
\end{equation}
This also enables us to neglect the inferred persona sentences that are incorrect or irrelevant to the dialogue history. 
Since we cannot guarantee that inferred persona information is complete, we calculate the weighted sum of $\hat{\bm{H}}_D $, $\hat{\bm{H}}_P$ and $\bm{H}_D$ to obtain the final hidden states as the decoder's input as follows:
\begin{equation} \label{eq: final hidden state}
\begin{aligned}
    &\bm{H}_{final} = \lambda_1 \cdot \hat{\bm{H}}_D + \lambda_2 \cdot \hat{\bm{H}}_P + \lambda_3 \cdot \bm{H}_D \\
    &\lambda_i = \frac{e^{w_i}}{\sum_{j}e^{w_j}} (i, j \in \{1,2,3\}), 
\end{aligned}
\end{equation}
where $ w_1, w_2, w_3 $ are additional model parameters with the same initial value.
This ensures that the essence of the original dialogue context is largely preserved.

Similar to \cite{liu2021towards}, we use special tokens to represent strategies and append them in front of the corresponding sentences. Our training objective can be formalized as:
\begin{equation}
\begin{aligned}
    & \hat{r} = s \oplus r \\
    & \mathcal{L}=-\frac{1}{N}\sum_{t=1}^N\text{log}P(\hat{r}_t|d,p,\hat{r}_{<t})
\end{aligned}
\end{equation}
where s stands for the strategy, r for the response, and N is the length of $\hat{r}$.

\subsection{Strategy-based Controllable Generation} \label{decoding strategy}
Supporters' responses in the emotional support task are annotated based on several support strategies, which are essential for providing effective support \cite{liu2021towards}. 
For instance, the supporter may choose to ask a \textit{Question} or provide statements of \textit{Reaffirmation and Confirmation} depending on the situation.
We provide more descriptions of these strategies in Appendix \ref{appendix: helping strategies}. Accordingly, it becomes intuitive that selecting different strategies corresponds to the available knowledge of the users' persona, demonstrating the importance of strategy selection in our proposed approach.
For instance, supporters could choose \textit{Providing Suggestions} if they have sufficient knowledge of the user's persona and situation, while they would resort to \textit{Question} if they lack such information.
Therefore, we propose an innovative strategy-based controllable generation method for the decoding phase. 
We decompose the generation probability into
\begin{equation} \label{eq: controllable generation}
\begin{aligned}
    P_{final}(r_t|r_{<t}, d, p) \propto & P(r_t|r_{<t}, d, p) \cdot \\ 
    &(\frac{P(r_t|r_{<t}, d, p)}{P(r_t|r_{<t}, d)})^\alpha
\end{aligned}
\end{equation}
where $ \alpha $ is the hyperparameter associated with the strategy, and $ d $ and $ p $ represent the dialogue history and persona, respectively. 
Both $P(r_t|r_{<t}, d, p)$ and $P(r_t|r_{<t}, d)$ are calculated by our model; the only difference is that persona is not included in calculating $P(r_t|r_{<t}, d)$.
The last term in this equation can be interpreted as the ratio of the probability of a token whether the persona is entered or not.
As the ratio increases, the token becomes more relevant to persona information, increasing the likelihood of generating the token after adding such persona information. Therefore, employing Eq.\ref{eq: controllable generation} increases the likelihood of more relevant tokens to the persona information.
$\alpha$ is set to different values depending on the strategy. The values used by all strategies are listed in Table \ref{tab:alpha_values}.

We investigate the values of $ \alpha $ corresponding to different strategies and define three categories: high, medium, and low, which correspond to 0.75, 0.375, and 0, respectively. More details about the tuning process of these values are discussed in Appendix \ref{appendix: tuning of alpha}.

\begin{table}[!t]
\centering
\scalebox{0.9}{
\begin{tabularx}{0.5\textwidth}{lll}
\toprule
\textbf{Strategy} & $ \bm \alpha $ & \textbf{Category} \\ \midrule
Question & 0 & low \\
Restatement or Paraphrasing & 0.75 & high \\
Reflection of Feelings & 0 & low \\
Self-disclosure & 0 & low \\
Affirmation and Reassurance & 0.75 & high \\
Providing Suggestions & 0.75 & high \\
Information & 0.75 & high \\
Others & 0.375 & medium \\
\bottomrule
\end{tabularx}}
\caption{The values and levels of $ \alpha $ corresponding to different strategies. }
\label{tab:alpha_values}
\end{table}

We provide explanations for two of our decided $ \alpha $ values. 
For effective support, there are two types of questions (\textit{Question} strategy) that can be asked from the seeker \cite{ivey2013intentional}: open and closed. Therefore, we choose the low level to avoid overthinking persona information, resulting in fewer open questions.
We chose the high level for the \textit{Providing Suggestions} strategy, as we needed to focus more on the persona information to provide more appropriate and specific suggestions.
See Appendix \ref{appendix: alpha selection} for explanations regarding the $\alpha$ of other strategies.

%% file: sections/experiment.tex
\subsection{Persona Extractor Evaluation} \label{sec: persona extractor evaluation}
\paragraph{Human Evaluation}
To validate the effectiveness of our persona extractor model, we first manually reviewed several inferences and discovered that the main errors could be categorized as contradictions (i.e., personas contain factual errors) or hallucinations (i.e., personas contain unreasonable and irrelevant deductions from the conversation). An example of contradictions would be if the seeker mentions in the conversation that he is a man, but the inferred persona is "I am a woman".
Moreover, an instance of hallucination errors would be if the inferred persona is "I am a plumber" when the seeker has not mentioned their occupation. 
Then, we chose 100 samples at random and hired workers on Amazon Mechanical Turk (AMT) to annotate each sample with one of the following four options: Reasonable, Contradictory, Hallucinatory, or Others. 
In addition, if the option Others was chosen, we asked workers to elaborate on the error. 
The annotators considered 87.3\% of the inferred persona samples as Reasonable while marking 8\% and 4\% of the samples as Contradictory and Hallucinatory, respectively. 
Moreover, only 0.667\% of the samples were marked as Others. However, upon further analysis, we found that such samples could also be classified in one of the mentioned error categories (see Appendix \ref{appendix: guidelines} for more details). The inter-annotator agreement, measured by Fleiss's kappa, was 0.458, indicating moderate agreement.

%主实验
\begin{table*}[!t]
\centering
\resizebox{0.9\textwidth}{!}{
\begin{tabularx}{\textwidth}{lllllllllllll}
\toprule
\textbf{Model} & \textbf{ACC}$\uparrow$ & \textbf{PPL}$\downarrow$ & \textbf{B-2}$\uparrow$ & \textbf{B-4}$\uparrow$ & \textbf{D-1}$\uparrow$ & \textbf{D-2}$\uparrow$ & \textbf{E-1}$\uparrow$ & \textbf{E-2}$\uparrow$ & \textbf{R-L}$\uparrow$ & \textbf{Cos-Sim}$\uparrow$ \\ \midrule
Blenderbot-Joint & 27.72 & 18.11 & 5.57 & 1.93 & 3.74 & 20.66 & 4.23 & 20.28 & 16.36 & 0.184\\
MISC & 31.34 & 16.28 & 6.60 & 1.99 & 4.53 & 19.75 & 5.69 & 30.76 & 17.21 & 0.187\\
Hard Prompt & 34.24 & 17.06 & 7.57 & 2.53 & \textbf{5.15} & 25.47 & 6.02 & 31.64 & 18.12 & 0.199 \\ \midrule
PAL ($\alpha=0$) & 34.25 & \textbf{15.92} & \textbf{9.28} & \textbf{2.90} & 4.72 & 25.56 & 5.87 & 33.05 & \textbf{18.27} & 0.229 \\
PAL & \textbf{34.51} & \textbf{15.92} & 8.75 & 2.66 & 5.00 & \textbf{30.27} & \textbf{6.73} & \textbf{41.82} & 18.06 & \textbf{0.244} \\
\bottomrule
\end{tabularx}}
\caption{The results of automatic metrics evaluation for each model on ESConv. PAL ($\alpha=0$) represents setting the $\alpha$ of each strategy to 0, thus neglecting our proposed controllable generation decoding method.}
\label{tab: main auto eval results}
\end{table*}

\begin{table*}[htbp]
\centering
\resizebox{0.67\textwidth}{!}{
\begin{tabular}{lccccccccc}
\toprule
\multirow{2}{*}{\textbf{PAL vs.}} & \multicolumn{3}{c}{\textbf{Blenderbot-Joint}} & \multicolumn{3}{c}{\textbf{MISC}}           & \multicolumn{3}{c}{\textbf{PAL ($\alpha=0$)}}   \\
                                  & \textbf{Win}  & \textbf{Lose}  & \textbf{Draw} & \textbf{Win} & \textbf{Lose} & \textbf{Draw} & \textbf{Win} & \textbf{Lose} & \textbf{Draw} \\
\midrule
\textbf{Coherence}                & \textbf{68}$^\ddagger$   & 26             & 6            & \textbf{54}$^\dagger$  & 34            & 12           & 46           & \textbf{48}   & 6            \\
\textbf{Identification}           & 42            & \textbf{44}    & 14           & \textbf{46}  & 42            & 12           & \textbf{58}$^\ddagger$  & 32            & 10           \\
\textbf{Comforting}               & \textbf{50}$^\ddagger$   & 32             & 18           & \textbf{62}$^\ddagger$  & 24            & 14           & \textbf{44}  & 42            & 14           \\
\textbf{Suggestion}               & \textbf{54}$^\ddagger$   & 32             & 14           & \textbf{42}  & \textbf{42}   & 16           & \textbf{46}  & 38            & 16           \\
\textbf{Information}              & \textbf{44}$^\dagger$   & 34             & 22           & \textbf{62}$^\ddagger$  & 22            & 16           & \textbf{52}  & 44            & 4            \\
\midrule
\textbf{Overall}                  & \textbf{52}$^\ddagger$   & 16             & 32           & \textbf{44}$^\ddagger$  & 28            & 28           & \textbf{40}$^\ddagger$  & 28            & 32          
\\
\bottomrule
\end{tabular}}
\caption{The results of the human interaction evaluation (\%). PAL performs better than all other models (sign test, $\ddagger$ / $\dagger$ represent \textit{p}-value < 0.05 / 0.1).}
\label{tab: human ab test}
\end{table*}

\subsection{Baselines}
\paragraph{Blenderbot-Joint \textmd{\cite{liu2021towards}}:} Blenderbot \cite{roller2021recipes} fine-tuned on the ESConv dataset. This model is trained to predict the correct strategy for the next response via the language modeling objective. In addition, this model can also be seen as PAL trained without incorporating persona.
\paragraph{MISC \textmd{\cite{tu2022misc}}:} the state-of-the-art (SOTA) on the ESConv benchmark, which leverages commonsense reasoning to better understand the seeker's emotions and implements a mixture of strategies to craft more supportive responses.
\paragraph{Hard Prompt:} this model employs a straightforward idea when modeling seekers' persona information in the emotional support task, in which persona information is concatenated to the dialogue history. That is, the input to the model would be in the form "\textit{Persona: \{persona\} $\backslash n$ Dialogue history: \{context\} $\backslash n$ Response: }".

\subsection{Implementation Details}
We conducted the experiments on PESConv and use a 7:2:1 ratio to split this dataset into the train, validation, and test sets. As \citet{liu2021towards} stated, Blenderbot \cite{roller2021recipes} outperforms DialoGPT \cite{zhang2020dialogpt} in this task. Therefore, similar to previous work \cite{liu2021towards, tu2022misc}, we used the 90M version of Blenderbot \cite{roller2021recipes}.
Moreover, we used the AdamW \cite{loshchilov2018fixing} optimizer with $\beta_1=0.9$ and $\beta_2=0.999$. 
We initialized the learning rate as 2.5e-5 and performed a 100-step linear warmup. 
The training and validation batch sizes were set to 4 and 16, respectively. 
The model was trained for 10 epochs, and we chose the checkpoint with the lowest loss on the validation set. 
During the decoding phase, we used both Top-\textit{k} and Top-\textit{p} sampling with $k=10$, $p=0.9$, with temperature and the repetition penalty set to set to 0.5 and 1.03, respectively.
The experiments were run on a single Quadro RTX 6000 GPU using the transformers library\footnote{\url{https://github.com/huggingface/transformers}} \cite{wolf2020transformers}. 

\subsection{Automatic Evaluation}
We adopted strategy prediction accuracy (ACC), perplexity (PPL), BLEU-n (B-n) \cite{papineni2002bleu}, Distinct-n (D-n) \cite{li2016diversity}, EAD-n (E-n) \cite{liu2022rethinking}, Rouge-L (R-L) \cite{lin2004rouge}, and the mean of the cosine similarity between supporters' responses and personas using the SimCSE \cite{gao2021simcse} representation (cos-sim) to automatically evaluate our model's performance. In addition, since the responses in this task are often long, we also leveraged the Expectancy-Adjusted Distinct (EAD) score to evaluate response diversity as the Distinct score has been shown to be biased towards longer sentences \cite{liu2022rethinking}. 
To calculate this score, rather than dividing the number of unique n-grams by the total number of n-grams, as done in the original Distinct score, we would use the model's vocabulary size as the denominator.

As shown in Table \ref{tab: main auto eval results}, PAL outperforms all baselines in automatic metrics, including the current SOTA model MISC. As Blenderbot-Joint can be perceived as PAL without persona employed in training, the significance of persona can be demonstrated through the comparison of the results achieved by PAL and PAL ($\alpha=0$) with Blenderbot-Joint.
In addition, compared to PAL ($\alpha=0$), PAL demonstrates a more balanced performance and has the best strategy prediction accuracy, diversity, and better alignment with persona information, which indicates more seeker-specific responses. 
Interestingly, the cos-sim value for PAL is comparable to the mean value of the dialogues with an empathy score of 5 in Figure \ref{fig:simcse analysis_emp}.
Through further comparing the performance of PAL and PAL ($\alpha=0$), we can see that our strategy-based decoding approach significantly improves the dialogue diversity, as shown by D-n and E-n, which are more important metrics for dialogue systems than B-n and R-L \cite{liu2016not, gupta2019investigating, liu2022rethinking}.

\begin{figure}[!b]
\centering
\includegraphics[width=\linewidth]{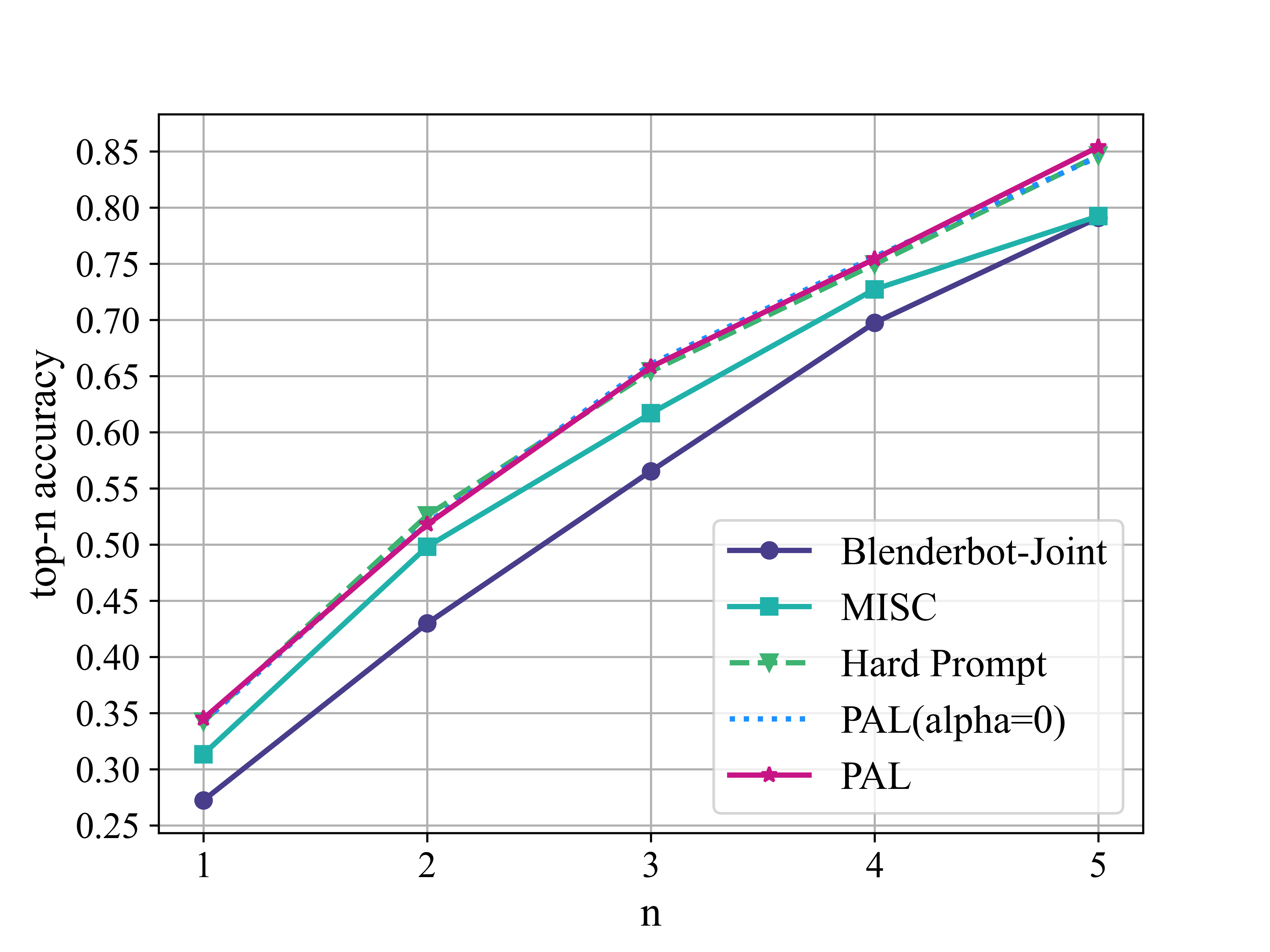} 
\caption{The top-n strategy prediction accuracy.}
\label{fig: strategy_acc} 
\end{figure}

\begin{table*}[!ht]
\begin{tabular}{|ll|}
\hline
\multicolumn{2}{|c|}{\textbf{Situation}}                                                                                                  \\ \hline
\multicolumn{1}{|l|}{Seeker}                  & I have just cheated on my girlfriend. I feel very guilty about it.                 \\ \hline
\multicolumn{2}{|c|}{\textbf{Dialogue history}}                                                                                           \\ \hline
\multicolumn{1}{|l|}{Seeker}                  & Hi, my friend.                                                                     \\ \hline
\multicolumn{1}{|l|}{Supporter}               & Hello ! How are you doing?                                                       \\ \hline
\multicolumn{1}{|l|}{Seeker}                  & Feeling very shame.                                                               \\ \hline
\multicolumn{1}{|l|}{ } & ......                                                                                                        \\ \hline
\multicolumn{1}{|l|}{\multirow{2}{*}{Seeker}} & But till now my girlfriend don't know about it. But her mom is now targeting me    \\ 
\multicolumn{1}{|l|}{}                        & for her sexual desire.                                                            \\ \hline
\multicolumn{2}{|c|}{\textbf{Persona Information}} \\ \hline
\multicolumn{1}{|l|}{Seeker}                  & I am feeling ashamed.                                 \\ \hline
\multicolumn{1}{|l|}{Seeker}                  & I have cheated on my girlfriend with her mother.                                 \\ \hline
\multicolumn{1}{|l|}{ } & ......                                                                                                       \\ \hline
\multicolumn{2}{|c|}{\textbf{Response}}                                                                                                   \\ \hline
\multicolumn{1}{|l|}{Blenderbot-Joint}        & I understand, I know how you feel. (\textit{Poor Empathy})                                                \\ \hline
\multicolumn{1}{|l|}{MISC}                    & I think you will be fine. (\textit{Poor Empathy})                                                        \\ \hline
\multicolumn{1}{|l|}{Hard Prompt}             & Oh no, I am so sorry, that is not good. (\textit{Poor Empathy})                                          \\ \hline
\multicolumn{1}{|l|}{PAL ($\alpha=0$)}           & I understand it is hard, so now you have to forgive her. (\textit{Less Proper Suggestion})   \\ \hline
\multicolumn{1}{|l|}{\multirow{3}{*}{PAL}}   & I understand how that can be hard. I would suggest you to talk to her mother,    \\  
\multicolumn{1}{|l|}{}                        & tell her that you \textbf{feel ashamed} about it and \textbf{don't cheat on your girlfriend again}. \\ 
\multicolumn{1}{|l|}{}                        & (\textit{Strong Empathy}) \\ \hline
\multicolumn{1}{|l|}{\textit{\textbf{Ground-truth}}}            & You have got such a nice girlfriend, have a happy life with her.                  \\ \hline
\end{tabular}
\caption{Responses from our approach and others. Due to space constraints, we have omitted some sentences.}
\label{case study}
\end{table*}

In Figure \ref{fig: strategy_acc}, we show the accuracy of the top-n strategy prediction results and our model PAL has the best results. It is worth noting that all models with persona information, PAL, PAL ($\alpha=0$), and Hard Prompt, all outperform MISC, demonstrating the importance of seekers' persona and highlighting the need for further research into how to better leverage such information in addition to commonsense reasoning.

\subsection{Human Evaluation}
We acknowledge that automatic metrics are insufficient for empirically evaluating and highlighting the improvements of our proposed method.
Hence, following \citet{liu2021towards}, we also conducted human evaluation by recruiting crowd-sourcing workers that interacted with the models. 
We provided workers with a scenario and asked them to act as seekers in those situations.
Each worker must interact with two different models and score them in terms of (1) Coherence; (2) Identification; (3) Comforting; (4) Suggestion; (5) Informativeness; and (6) Overall Preference. Detailed explanations for each aspect can be found in Appendix \ref{appendix: guidelines}.

As shown in Table \ref{tab: human ab test}, we compare PAL with the other three models, and PAL beats or is competitive with other methods on all of the above metrics. It performs well on three key metrics more closely aligned with persona (i.e., Comforting, Suggestion, and Information), implying that persona is required in emotional support.

%% file: sections/case_study.tex
%case study的表格在experiment里
In Table \ref{case study}, we provide an example to compare the responses of our approach with the other methods. As can be seen, the Blenderbot-Joint, MISC, and Hard Prompt methods all provide only very poor empathy, with responses that are very general and do not contain much information. Whereas PAL ($\alpha=0$), which does not use the strategy-based decoding method, is more specific but provides a less appropriate suggestion. 
Our model PAL shows strong empathy, is the most specific while providing appropriate suggestions, and incorporates persona information in the response (\textit{feel ashamed} and \textit{don't cheat on your girlfriend again}). Due to space constraints, more cases, including cases of interactions and analysis over different strategies, can be found in Appendix \ref{appendix: case study}.

%% file: sections/conclusion.tex
In this work, we introduced persona information into the emotional support task. We proposed a framework that can dynamically capture seekers' persona information, infer persona information using our trained persona extractor, and generate responses with a strategy-based controllable generation method. Through extensive experiments, we demonstrated that our proposed approach outperformed the studied baselines in both human and manual evaluation. In addition, we provided persona annotations for the ESConv dataset using the persona extractor model, which will foster the research of personalized emotional support conversations.

%% file: sections/limitations.tex
\paragraph{Persona extractor} First, we need to clarify that our definition of persona is not exactly psychological, the role an individual plays in life \cite{CGJung2014PsychologicalT}. As a result, like previous studies (e.g., Persona-Chat \cite{zhang2018personalizing}, PEC \cite{zhong2020towards}), the format of persona is flexible and variable. As stated in \S\ref{sec: persona extractor evaluation}, there are still some issues with the model we use to infer persona information. For example, we sometimes get information that contradicts the facts. And also, there is occasionally unrelated content, as with commonsense 
reasoning \cite{tu2022misc}. Furthermore, we cannot guarantee that we can infer all of the persona information that appears in the conversation because much of it is frequently obscure. And when extracting persona information, we only use what the user said previously and remove what the bot said, which results in the loss of some conversation information. The reason for this is that we have discovered that if we use the entire conversation, the model frequently has difficulty distinguishing which persona information belongs to the user and which belongs to the other party.
In addition, since the code of \citet{xu2022long} is not yet available, we have not compared other methods of extracting persona dynamically from the conversation.

\paragraph{Strategy-based decoding} During the decoding phase, we only coarse-grained the $\alpha$ of each strategy because we discovered that only coarse-grained tuning produced good results, and future work may be able to further explore the deeper relationship between different strategies and persona.

%% file: sections/ethics.tex
In this work, we leveraged two publicly available datasets.
First, we used the Persona-Chat dataset, which is collected by assigning a set of fixed pre-defined persona sentences to workers.
Therefore, by participating in this dataset, workers were required not to disclose any personal information \cite{zhang2018personalizing}, which prevents issues regarding the leakage of their privacy.
Similarly, during the collection of the ESConv dataset, participants were asked to create imaginary situations and play the role of a support seeker who is in that situation. 
In addition, they were instructed not to provide personal information during their conversations with the trained supporters \cite{liu2021towards}.
Regarding the persona extractor, this module is trained to infer and extract persona information solely from what the user has mentioned in the conversation rather than making assumptions about the user's background and character, further highlighting the importance of user privacy in our research.

Regarding our experiments, we ensured that all workers agreed to participate in the annotation tasks.
Moreover, as the workers were recruited from the US, we ensured that they were paid above the minimum wage in this country for successfully completing our tasks.
We acknowledge that using trained dialogue models to provide support is a sensitive subject and research on this topic should be conducted with sufficient precautions and supervision.
We also acknowledge that in their current stage, such models cannot replace human supporters for this task \cite{Emohaa}. Thus, they should not be employed to replace professional counselors and intervention and interact with users that suffer from mental distress, such as depression or suicidal thoughts.

%% file: sections/appendix.tex
\section{Persona Extractor} \label{appendix: persona extractor}
In our initial experiments, we compare the effectiveness of various generative models to infer persona (such as GPT2 \cite{radford2019language}, DialoGPT \cite{zhang2020dialogpt}, BART \cite{lewis2020bart}). We manually checked some results and found the best results were obtained by the Bart model fine-tuned on CNN Daily Mail \cite{nips15_hermann}. We trained this model for ten epochs with a batch size of 4 and learning rate of 1e-5, and selected the best-performing checkpoint.

\section{Relevance of Individualization and Seeker
Evaluation} \label{appendix: analysis}
\begin{figure*}[htbp]
\centering
\subfigure[Empathy]{
\begin{minipage}[t]{0.3\linewidth}
\centering
\includegraphics[width=2in, height=1.5in]{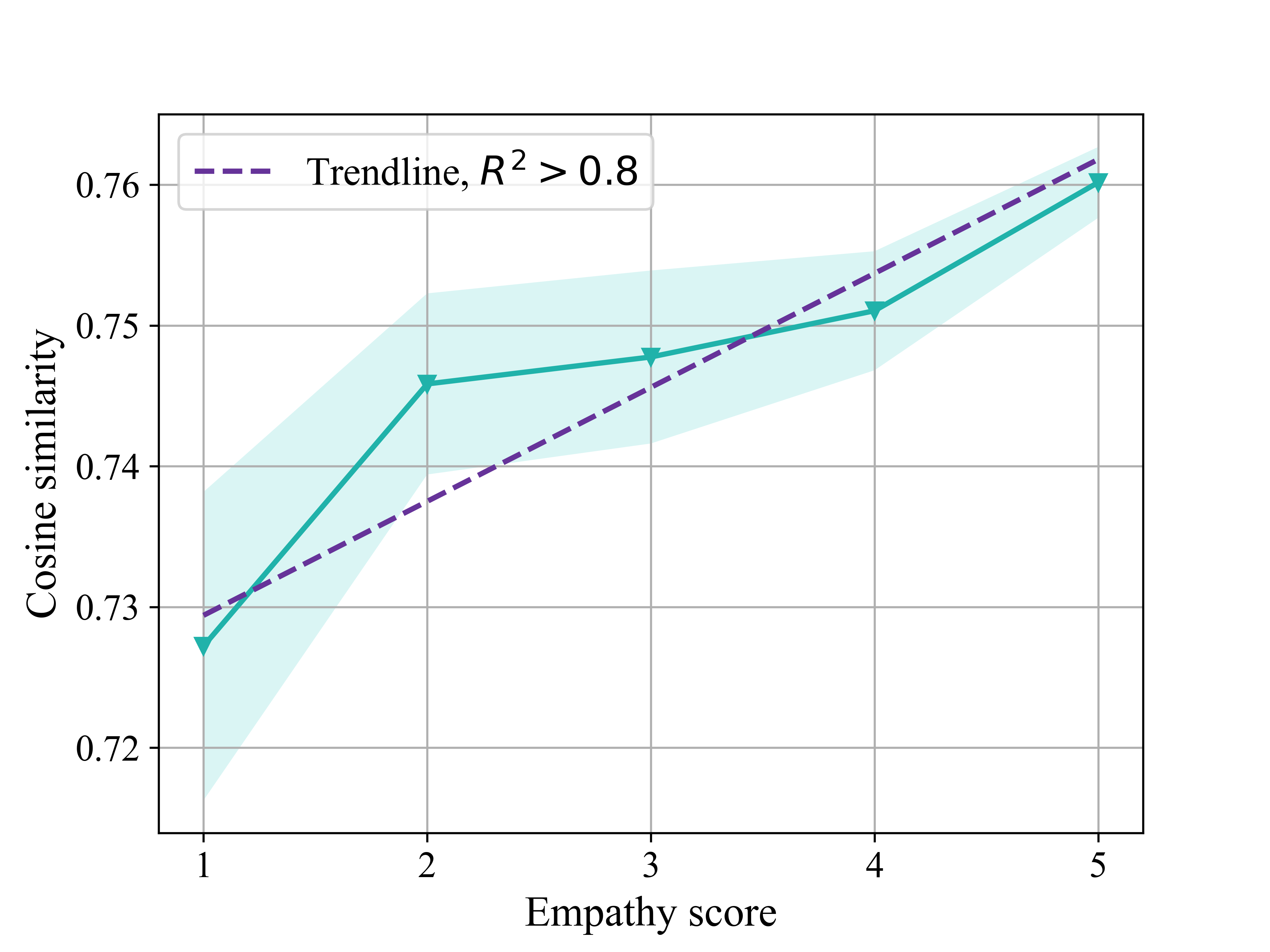}
\end{minipage}
}
\subfigure[Relevance]{
\begin{minipage}[t]{0.3\linewidth}
\centering
\includegraphics[width=2in, height=1.5in]{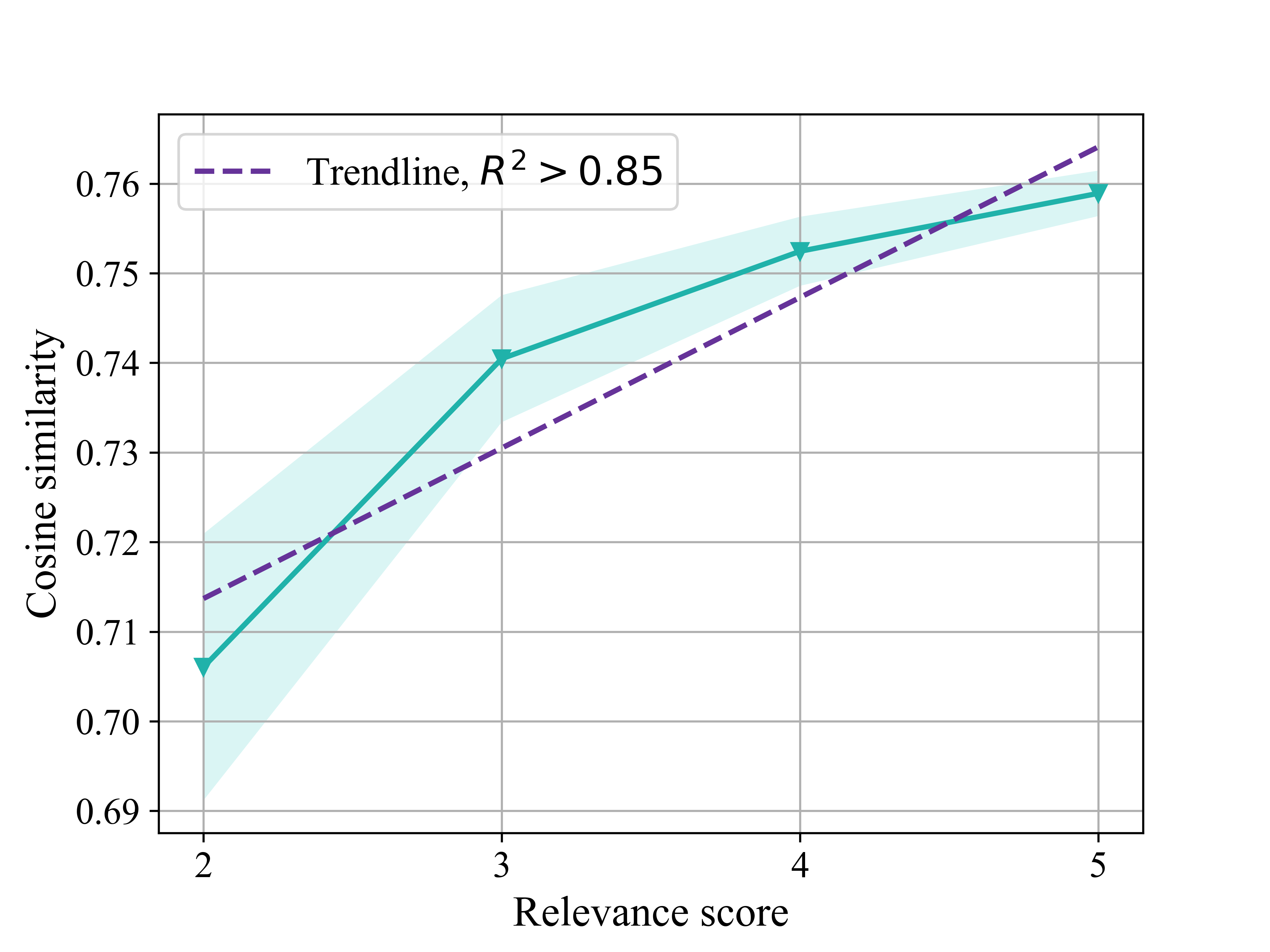}
\end{minipage}
}
\subfigure[Decrease in Emotional Intensity]{
\begin{minipage}[t]{0.3\linewidth}
\centering
\includegraphics[width=2in, height=1.5in]{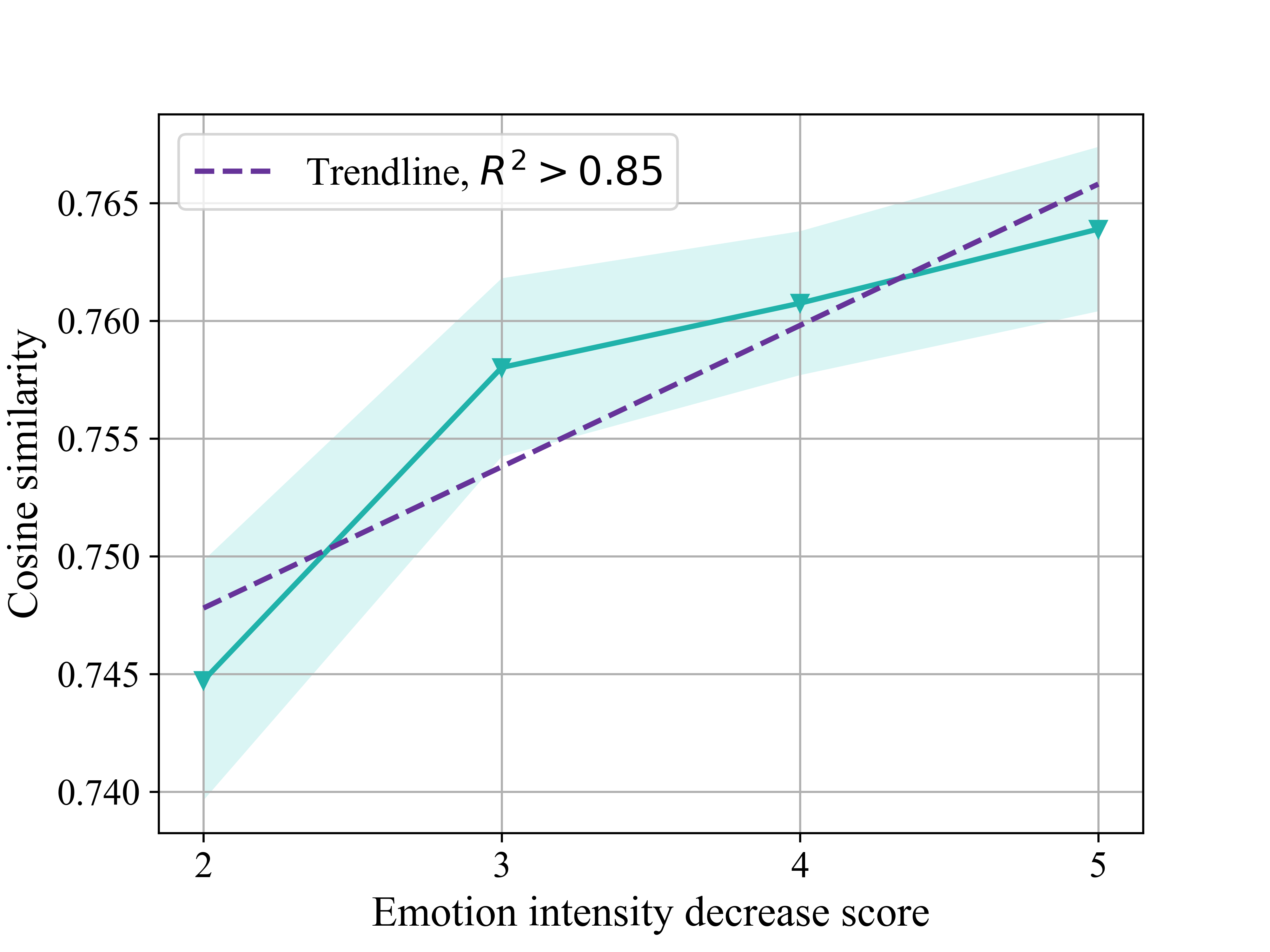}
\end{minipage}
}
\centering
\caption{The relationship between the empathy score, relevance score, emotion intensity decrease score, and the similarity between supporters' responses and persona information using fastText. We can observe that in general, more similarity leads to higher scores. In addition, we display the trend line and the coefficient of determination.}
\label{fig: fasttext analysis}
\end{figure*}

Here we show the results produced by fastText in Figure \ref{fig: fasttext analysis}.

\section{Helping Strategies in ESConv} \label{appendix: helping strategies}
A total of 8 strategies are marked in ESConv, and they are basically evenly distributed \cite{liu2021towards}.
Here we list these strategies and their detailed definitions, which are directly adopted from \citet{liu2021towards}.
\paragraph{Question} Asking for information related to the problem to help the help-seeker articulate the issues that they face. Open-ended questions are best, and closed questions can be used to get specific information.
\paragraph{Restatement or Paraphrasing} A simple, more concise rephrasing of the help-seekers' statements could help them see their situation more clearly.
\paragraph{Reflection of Feelings} Articulate and describe the help-seekers' feelings.
\paragraph{Self-disclosure} Divulge similar experiences that you have had or emotions that you share with the help-seeker to express your empathy.
\paragraph{Affirmation and Reassurance} Affirm the help-seeker's strengths, motivation, and capabilities and provide reassurance and encouragement.
\paragraph{Providing Suggestions} Provide suggestions about how to change but be careful not to overstep and tell them what to do.
\paragraph{Information} Provide useful information to the help-seeker, for example, with data, facts, opinions, resources, or by answering questions.
\paragraph{Others} Exchange pleasantries and use other support strategies that do not fall into the above categories.

\section{Tuning Process of the $\alpha$ Values} \label{appendix: tuning of alpha}
We first tried to set these alpha values as trainable parameters, but we found that the values changed very little during the training of the model and therefore depended heavily on the initialization, so we set these alpha's as hyperparameters. Then, these values were obtained upon numerous attempts on the validation set as they enabled the model to have a balanced performance based on the automatic evaluation. 
We acknowledge that this tuning process is trivial and coarse-grained. We leave approaches to improve this process, such as using a simulated annealing algorithm, to future work.

\section{Analysis of $\alpha$ Selected for Different Strategies} \label{appendix: alpha selection}
In \S\ref{decoding strategy}, we analyzed the strategies \textit{Question} and \textit{Providing Suggestions}. And the rest of the strategies are analyzed below.

For the \textit{Restatement or Paraphrasing} strategy, it is necessary to repeat the words of the seeker, so a more specific restatement can help the seeker better understand himself.
For the \textit{Reflection of Feelings} strategy, since the focus is more on feelings, and the extracted persona information is more fact-related, we set low for this strategy.
For the \textit{Self-disclosure} strategy, it is more about the supporter's own experience and should not focus too much on the persona information of the seeker, which may lead to unnecessary errors, so we set this strategy to low.
For the \textit{Affirmation and Reassurance} strategy, combining the seeker's persona information can often provide more specific encouragement and bring the seeker a better experience, so we set it to high.
For the \textit{Information} strategy, we need to consider more persona information in order to provide more appropriate and specific information for seekers, so we set it high.
For the \textit{Other} strategy, the main places this appear are greeting and thanking. About this strategy, considering that most appearances are in greeting and thanking, if we can combine more seeker characteristics may make seekers feel more relaxed, we set it to the high level at first, but careful observation found that \textit{Other} strategies are used when the other strategies are not appropriate. Although such cases are rare, in order to avoid unnecessary errors, we set it to medium.

\section{Human Evaluation} \label{appendix: guidelines}
\begin{figure*}[htbp]
\centering
\includegraphics[width=\linewidth]{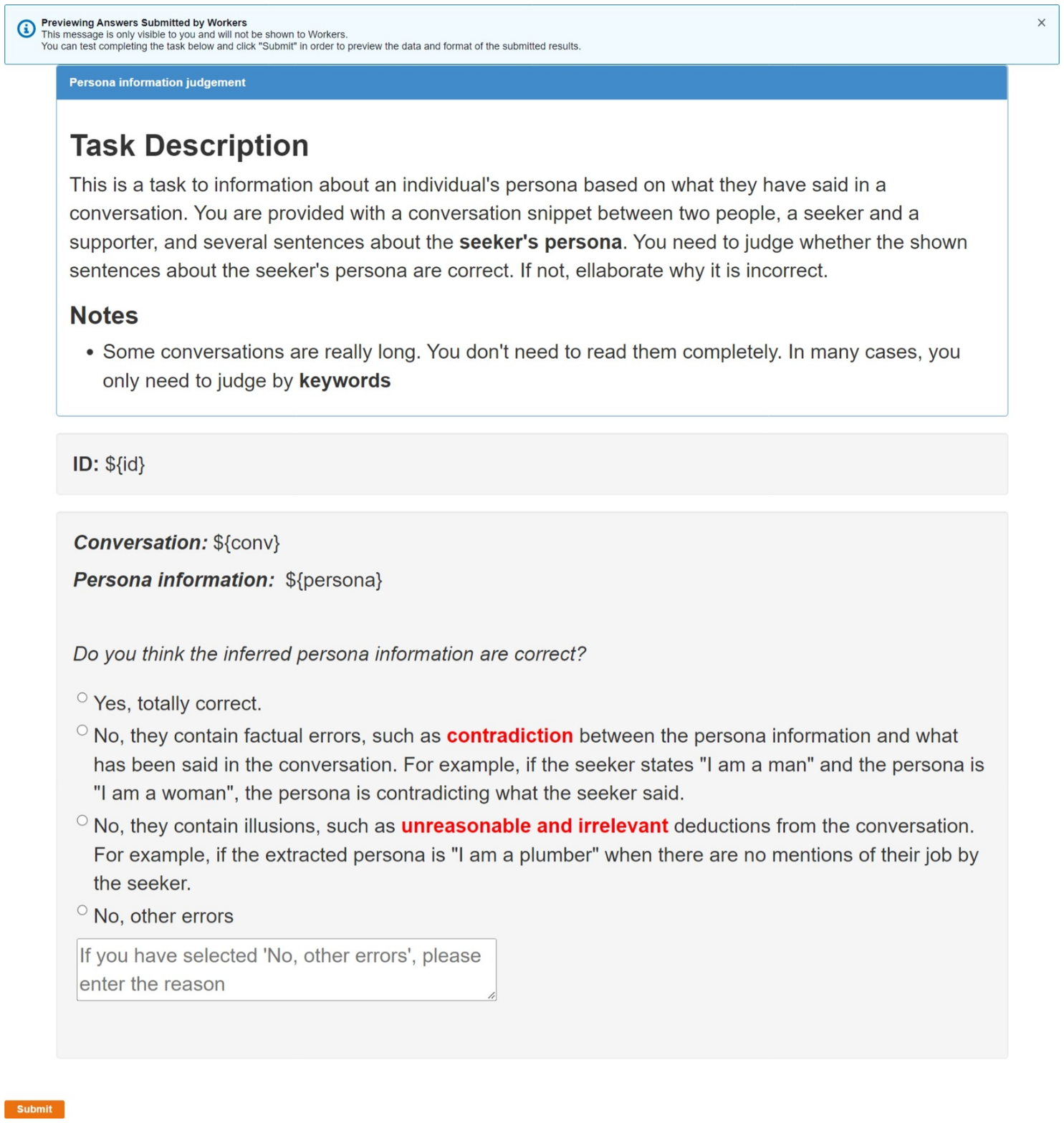} 
\caption{The screenshot of persona extractor human evaluation guideline.}
\label{fig: persona eval guideline} 
\end{figure*}

\begin{figure*}[htbp]
\centering
\includegraphics[width=\linewidth]{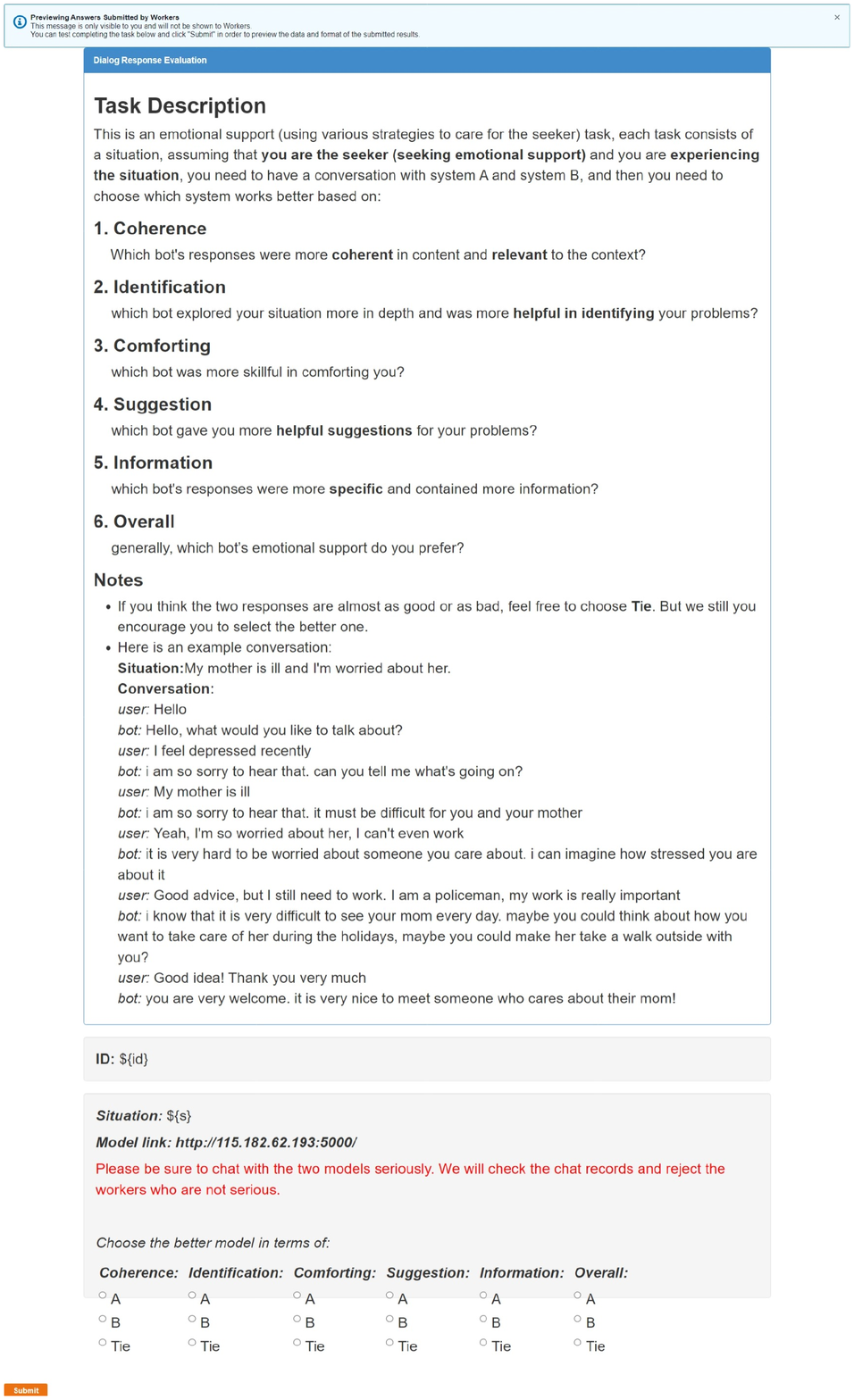} 
\caption{The screenshot of human interaction evaluation guideline.}
\label{fig: abtest guideline} 
\end{figure*}

Here we show the guidelines for two human evaluation experiments in Figure \ref{fig: persona eval guideline} and Figure \ref{fig: abtest guideline}. For the persona extractor manual evaluation experiment, we pay \$0.05 for one piece of data, and for the human interactive evaluation, we pay \$0.10 for one piece of data, with the price adjusted for the average time it takes workers to complete the task. We stated in the task description that this is an evaluation task, so for the data submitted by the workers, we only use it for evaluations.

\section{Case Study} \label{appendix: case study}
Due to space limitations, we show more examples here, these are cherry-picked.

\begin{figure*}[htbp]
\centering
\includegraphics[width=\linewidth]{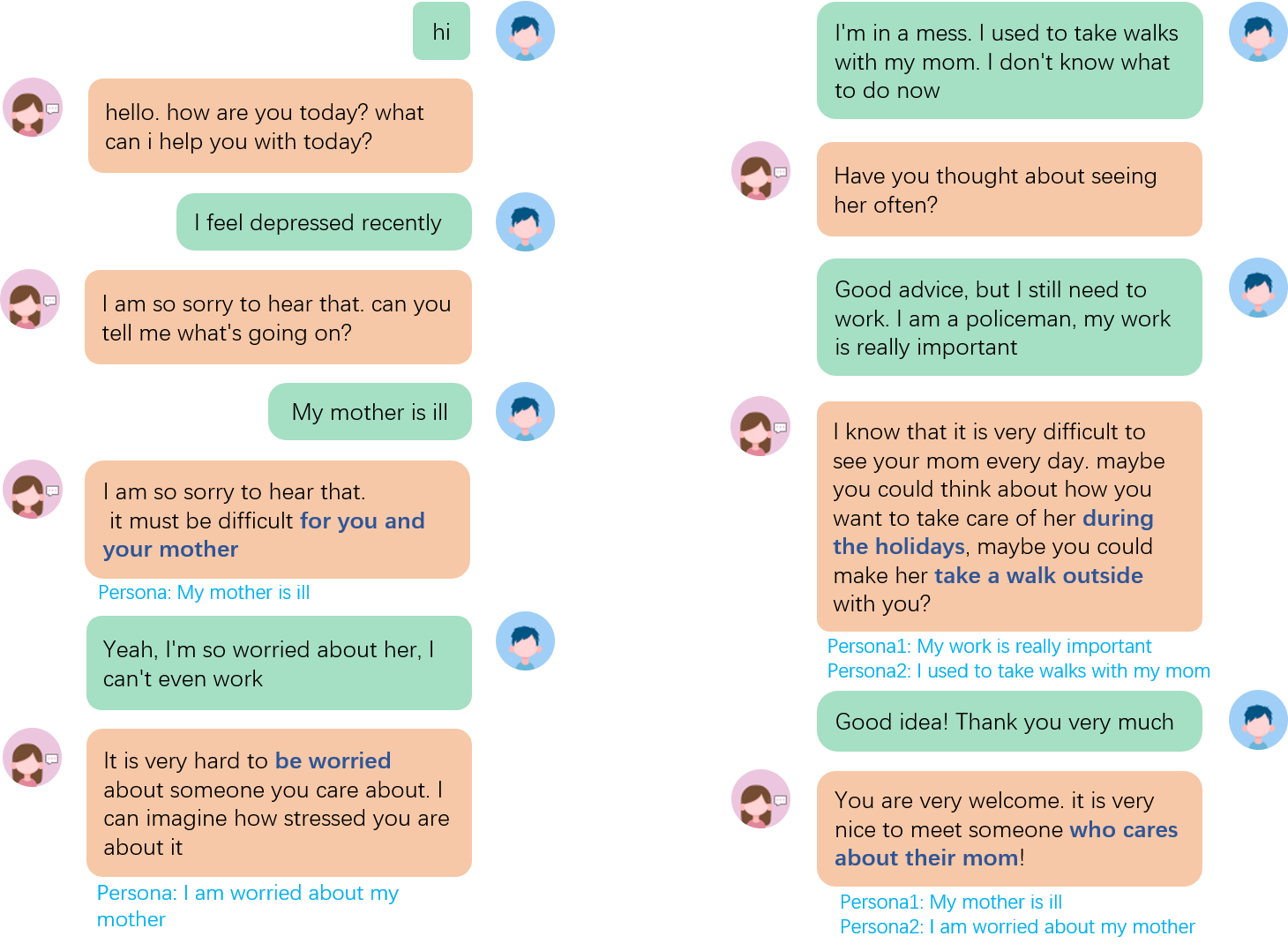} 
\caption{An interactive example. In this case, the orange dialog box is the supporter (PAL) and the green dialog box is the seeker. Due to the long dialogue rounds, it is split into left and right sides, with the dialogue on the left side going first. The persona used by PAL is given below the dialog box.}
\label{fig: interactive dialogue case} 
\end{figure*}

\begin{figure*}[!ht]
\centering
\includegraphics[width=\linewidth]{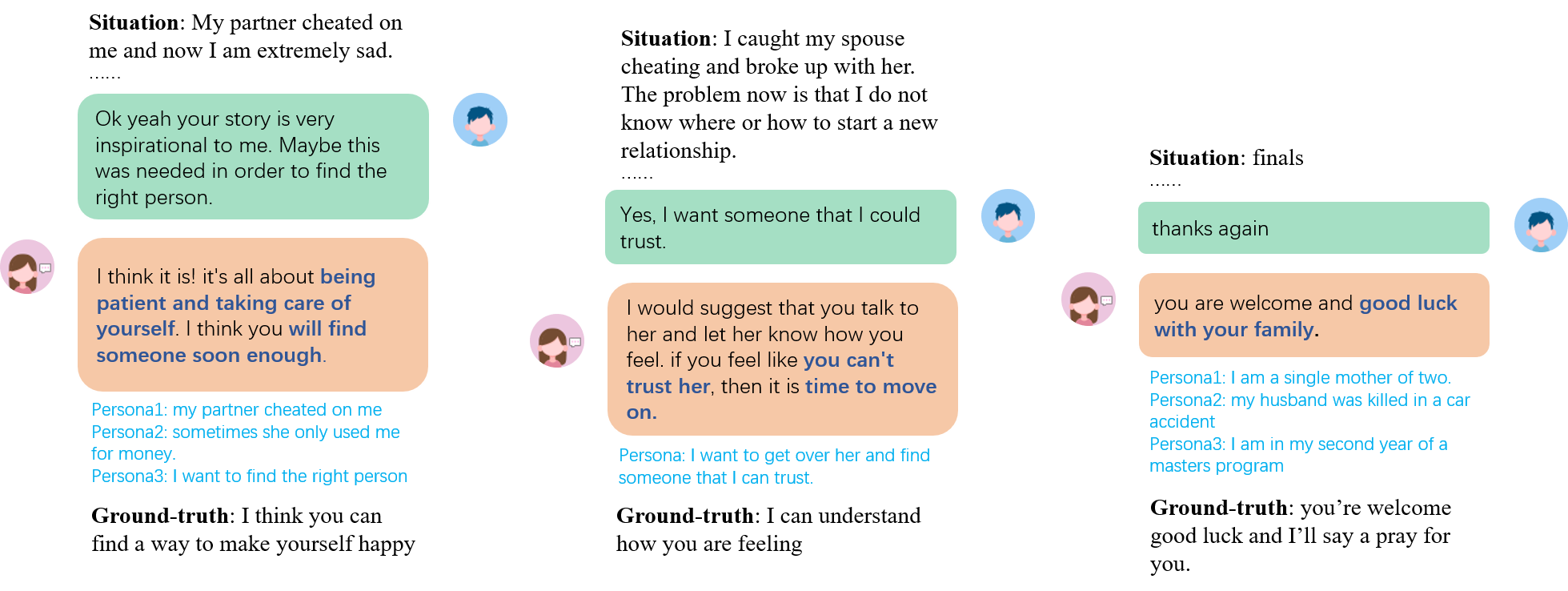} 
\caption{Some examples in the ESConv dataset where we do not show all rounds of dialogues due to space constraints. The orange dialogues are from supporters (PAL).}
\label{fig: esc case} 
\end{figure*}

In Figure \ref{fig: interactive dialogue case}, we show an interactive case. It can be seen that PAL uses the extracted persona appropriately several times in the conversation and gives the seeker specific advice.

In Figure \ref{fig: esc case}, we show some cases in the ESConv dataset. Interestingly, in these examples, PAL sometimes performs better than Ground-truth, giving a more appropriate and specific response rather than a general one, which also proves the superiority of our model.

% TODO cases over different strategies
Here, we also compare our model with baselines over different strategies. In Table \ref{appendix: providing suggestions}, we show a case of the strategy \textit{Providing Suggestions}. We can find that our model provides the most specific suggestions. In Table \ref{appendix: Affirmation and Reassurance}, we show a case of the strategy \textit{Affirmation and Reassurance}. We can also see that PAL's response is the most specific.

\begin{table*}[htbp]
\begin{tabular}{|ll|}
\hline
\multicolumn{2}{|c|}{\textbf{Situation}}                                                                                                  \\ \hline
\multicolumn{1}{|l|}{\multirow{2}{*}{Seeker}}   & My two friends are shaming me for not donating more per year to charities.    \\  
\multicolumn{1}{|l|}{}                        & One of them is the President of a local non-profit. \\ \hline
\multicolumn{2}{|c|}{\textbf{Dialogue history}}                                                                                           \\ \hline
\multicolumn{1}{|l|}{ } & ......                                                                                                        \\ \hline
\multicolumn{1}{|l|}{Seeker}                  & What should I do about these two (so-called) friends?                                                                     \\ \hline
\multicolumn{2}{|c|}{\textbf{Persona Information}} \\ \hline
\multicolumn{1}{|l|}{Seeker}                  & My friends are putting me on a guilt trip and shaming me.                                 \\ \hline
\multicolumn{1}{|l|}{Seeker} & I am not making much income with covid19.                                                                                                       \\ \hline
\multicolumn{2}{|c|}{\textbf{Response}}                                                                                                   \\ \hline
\multicolumn{1}{|l|}{Blenderbot-Joint}        & Well, they are not worth the friendship.                                                \\ \hline
\multicolumn{1}{|l|}{MISC}                    & Do you think that you can talk to them about how you feel?                                                        \\ \hline
\multicolumn{1}{|l|}{Hard Prompt}             & Maybe you should try to talk to them about it and see what they say about it.                                           \\ \hline
\multicolumn{1}{|l|}{PAL($\alpha=0$)}           & If you are hurt because of them then I would suggest you seek professional help.   \\ \hline
\multicolumn{1}{|l|}{\multirow{2}{*}{PAL}}   & If they are {\color{red} hurting you physically} then I would suggest that you find a way to talk     \\  
\multicolumn{1}{|l|}{}                        & to them  and {\color{red} explain your income situation}. \\ \hline
\multicolumn{1}{|l|}{\textit{\textbf{Ground-truth}}}            & Definitely not  ...have you tried explaining your situation to them?                  \\ \hline
\end{tabular}
\caption{Responses from our approach and others.}
\label{appendix: providing suggestions}
\end{table*}

\begin{table*}[htbp]
\begin{tabular}{|ll|}
\hline
\multicolumn{2}{|c|}{\textbf{Situation}}                                                                                                  \\ \hline
\multicolumn{1}{|l|}{Seeker}   & I caught my spouse cheating and broke up with her. \\ \hline
\multicolumn{2}{|c|}{\textbf{Dialogue history}}                                                                                           \\ \hline
\multicolumn{1}{|l|}{ } & ......                                                                                                        \\ \hline
\multicolumn{1}{|l|}{Seeker}                  & She got a job in another town and she utilize the opportunity to be sleeping around.                          \\ \hline
\multicolumn{2}{|c|}{\textbf{Persona Information}} \\ \hline
\multicolumn{1}{|l|}{Seeker}                  & My wife is cheating on me.                  \\ \hline
\multicolumn{1}{|l|}{} & ......                                                                                                       \\ \hline
\multicolumn{2}{|c|}{\textbf{Response}}                                                                                                   \\ \hline
\multicolumn{1}{|l|}{Blenderbot-Joint}        & Wow... I am sorry to hear that.                                                \\ \hline
\multicolumn{1}{|l|}{MISC}                    & I understand.. I had a similar situation in a previous relationship.                                                        \\ \hline
\multicolumn{1}{|l|}{Hard Prompt}             & So you trust your spouse but you have disappointed your marriage vows.                                           \\ \hline
\multicolumn{1}{|l|}{PAL($\alpha=0$)}           & I can see how that can be a problem. That is a very difficult situation to be in.  \\ \hline
\multicolumn{1}{|l|}{\multirow{2}{*}{PAL}}   & That's a difficult situation to be in. It sounds like you are {\color{red} being betrayed}.      \\  
\multicolumn{1}{|l|}{}                        &  I believe you deserve someone better. \\ \hline
\multicolumn{1}{|l|}{\multirow{2}{*}{\textit{\textbf{Ground-truth}}}}            & I can imagine how a break in trust has made you feel. I know a break in trust \\ 
\multicolumn{1}{|l|}{} & is always very difficult to process.                  \\ \hline
\end{tabular}
\caption{Responses from our approach and others.}
\label{appendix: Affirmation and Reassurance}
\end{table*}